\documentclass{article}

\usepackage{PRIMEarxiv}

\usepackage[utf8]{inputenc} 
\usepackage[T1]{fontenc}    
\usepackage{hyperref}       
\usepackage{url}            
\usepackage{booktabs}       
\usepackage{amsfonts}       
\usepackage{nicefrac}       
\usepackage{microtype}      
\usepackage{lipsum}
\usepackage{graphicx}       
\usepackage{multicol}
\usepackage{float}
\usepackage[none]{hyphenat}
\usepackage{enumitem}
\graphicspath{{media/}}     

\title{A Large-Scale, Physically-Based Synthetic Dataset for Satellite Pose Estimation
}

\author{
  Szabolcs Velkei \\
  Machine Intelligence Zrt \\
  Szigetmonostor, Hungary \\
  \texttt{szabolcs.velkei@mi.services} \\
   \And
  Csaba Goldschmidt \\
  Machine Intelligence Zrt \\
  Szigetmonostor, Hungary \\
  \texttt{csaba.goldschmidt@mi.services} \\
   \And
  K\'aroly Vass \\
  Machine Intelligence Zrt \\
  Szigetmonostor, Hungary \\
  \texttt{karoly.vass@mi.services} \\
}

\begin{document}
\maketitle

\begin{abstract}
The Deep Learning Visual Space Simulation System (DLVS\textsuperscript3) introduces a novel synthetic dataset generator and a simulation pipeline specifically designed for training and testing satellite pose estimation solutions. This work introduces the DLVS\textsuperscript3-HST-V1 dataset, which focuses on the Hubble Space Telescope (HST) as a complex, articulated target. The dataset is generated using advanced real-time and offline rendering technologies, integrating high-fidelity 3D models, dynamic lighting (including secondary sources like Earth reflection), and physically accurate material properties. The pipeline supports the creation of large-scale, richly annotated image sets with ground-truth 6-DoF pose and keypoint data, semantic segmentation, depth, and normal maps. This enables the training and benchmarking of deep learning-based pose estimation solutions under realistic, diverse, and challenging visual conditions. The paper details the dataset generation process, the simulation architecture, and the integration with deep learning frameworks, and positions DLVS\textsuperscript3 as a significant step toward closing the domain gap\cite{gallet2024exploring,bravo2022spacecraft,jawaid2022towards} for autonomous spacecraft operations in proximity and servicing missions.
\end{abstract}

\begin{multicols}{2}
\section{Introduction}
Accurate 6-Degrees-of-Freedom (6-DoF) pose estimation of satellites is a critical enabler for autonomous spacecraft operations, including rendezvous, docking, debris removal, and on-orbit servicing. Traditional approaches to pose estimation are hampered by the scarcity of labeled real-world data and the complexity of space environments, especially for non-cooperative and articulated targets. To address these challenges, the DLVS\textsuperscript3 project was initiated, aiming to generate high-fidelity synthetic datasets and provide a versatile simulation testbed for training and evaluating deep learning models in realistic orbital scenarios.

The DLVS\textsuperscript3 framework leverages state-of-the-art 3D modeling, physically-based rendering, and procedural environment generation to produce large-scale datasets with precise ground-truth annotations. The system is capable of simulating a wide range of lighting conditions, satellite articulations, and sensor effects, ensuring that the resulting data closely mirrors the operational realities of space missions. By focusing on the HST, one of the most visually documented and structurally complex satellites, the project demonstrates the feasibility and advantages of synthetic data-driven approaches for advancing pose estimation capabilities in the space domain.

\section{Related work}
The field of satellite pose estimation has seen significant advances with the advent of deep learning, but progress has been constrained by the limited availability and diversity of real-world datasets. Benchmark datasets such as TANGO\cite{kisantal2020satellite}, SPEED\cite{chen2019satellite}, SPEED+\cite{park2022speed}, SHIRT,\cite{Park2023AdaptiveNNUKF}, SPADES\cite{rathinam2023spades}, SPARK\cite{boukhtache2021spark}, SEENIC\cite{jawaid2022towards}, URSO\cite{proenca2019deep} have provided foundational resources, yet their scale (typically 50,000–100,000 images) and coverage of environmental and articulation variability remain insufficient for training robust models suited for real mission conditions. Existing simulators like SISPO\cite{iakubivskyi2022sispo}, SurRender\cite{delattre2021surrender}, PANGU\cite{sanchezgestido2024pangu} and SPIN \cite{montalvo2024spin} excel in specific applications but they present several critical limitations when used for generating training datasets for deep learning in computer vision and pose estimation tasks:
\begin{itemize}
\item These simulators typically support only a small number of light sources, often focusing on direct sunlight and simple shadow casting.
\item They lack the ability to procedurally generate a wide variety of planetary surfaces, cloud morphologies, atmospheric phenomena, and transient effects
\item Use basic, non-parameterized material models and do not account for the wide range of surface properties, procedural variation in reflectance, roughness, or generate random wrinkling for MLI like surfaces.
\item The output from these simulators is often restricted to rendered RGB images and basic pose information.
\end{itemize}
Recent research emphasizes the importance of large-scale, high-fidelity synthetic data to bridge the domain gap between simulation and reality. Techniques such as domain randomization, procedural material generation, material aging, environment modeling, and the inclusion of secondary illumination sources (e.g., Earthshine) have been shown to improve generalization and robustness in pose estimation networks. DLVS\textsuperscript3 builds upon these insights by integrating advanced rendering pipelines (using Houdini\cite{sidefxhoudini} and Unreal Engine\cite{epicgames_unrealengine}), physically accurate material libraries (MaterialX\cite{materialx}), and comprehensive annotation schemas (including keypoints, segmentation, depth, and normals) to generate datasets orders of magnitude larger and more diverse than previous efforts.

Furthermore, DLVS\textsuperscript3 distinguishes itself by supporting articulated satellite models, procedural animation, and real-time simulation control, enabling the generation of training and evaluation data tailored to the specific requirements of autonomous proximity operations. This positions the DLVS\textsuperscript3 HST dataset as a new benchmark for the development and assessment of deep learning-based satellite pose estimation solutions.

\section{Dataset generation methods}
\subsection{Domain randomization}
 DLVS\textsuperscript3 employs extensive domain randomization during dataset generation. This includes randomizing satellite poses, articulation states, lighting directions, camera parameters, and environmental conditions for each image. By systematically varying these factors, the generated data expose neural networks to a wide range of possible real-world scenarios, reducing overfitting and enhancing model robustness when deployed in operational settings. The randomization extends to both the position of primary (Sun) and secondary (Earth) light sources, as well as to the background environment, which is procedurally generated with high-resolution textures and dynamic atmospheric effects.
\subsection{Procedural materials and material aging}
A core innovation of  DLVS\textsuperscript3 is its use of a custom MaterialX-based material library, which enables highly realistic, physically-based rendering of satellite surfaces and components. Materials are not static; instead, their properties - such as reflectivity, roughness, anisotropy, and color - are procedurally randomized within physically plausible bounds for each sample. The simulation of material aging is a unique feature:  DLVS\textsuperscript3 models the effects of prolonged space exposure, including UV-induced discoloration, metal tarnishing, clearcoat degradation, and micrometeorite-induced surface damage. This is achieved through a combination of procedural texture generation, digital sculpting, and parameterized noise functions, allowing for the creation of virtually limitless material variations that mirror the diversity observed in real space missions.
\subsection{Environment Modeling}
\begin{figure}[H]
    \centering
    \includegraphics[width=0.9\columnwidth]{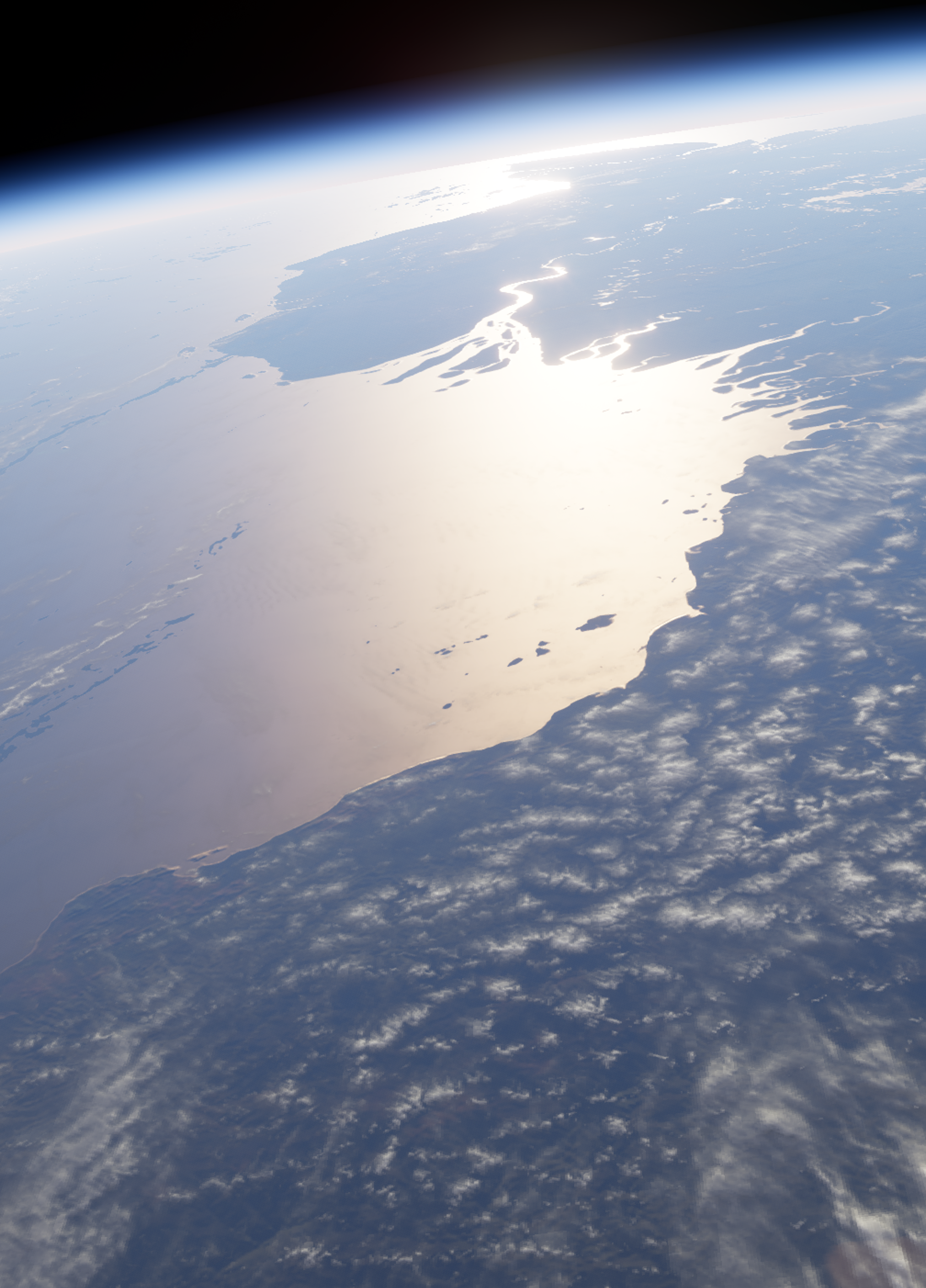}
    \caption{Atmospheric effects over Fly River's delta}
    \label{fig:fig1}
\end{figure}

 DLVS\textsuperscript3’s environment modeling leverages real-time (Unreal Engine) rendering pipelines to generate photorealistic planetary backgrounds. The Earth is modeled with scientifically accurate geometry and multi-layered, dynamic cloud systems, including atmospheric effects such as Rayleigh and Mie scattering, auroras, and city lights as seen in Figure \ref{fig:fig1}. These backgrounds are not merely visual; they serve as active illumination sources, contributing to the realism of reflections and shadows on the satellite surfaces. The system supports the procedural placement and movement of these planetary bodies, further increasing the diversity and realism of the training data.

\subsection{SPICE integration}
 DLVS\textsuperscript3 integrates NASA/NAIF's SPICE\cite{acton1996ancillary} (Spacecraft Planet Instrument Camera-matrix Events) toolkit to ensure astronomically accurate positioning and motion of celestial bodies in simulations. This implementation enables rigorous modeling of orbital dynamics, coordinate system transformations, and timekeeping, critical for high-fidelity space environment simulations.
\subsection{Illumination sources}
A major advancement in  DLVS\textsuperscript3 is the explicit modeling of secondary illumination sources, most notably Earthshine. In low Earth orbit, the Earth can act as a powerful diffuse light source, significantly altering the appearance of satellites in shadow or partially illuminated conditions.  DLVS\textsuperscript3 simulates these effects by rendering a 360-degree 32-bit floating high dynamic range (HDR) environment map in Unreal Engine that serves as an emissive dome surrounding the satellite during the Houdini ray-tracing rendering process. This approach ensures that the synthetic images capture lighting phenomena that are critical for robust pose estimation but are neglected in previous simulation frameworks.

 \begin{figure}[H]
    \centering
    \includegraphics[width=0.9\columnwidth]{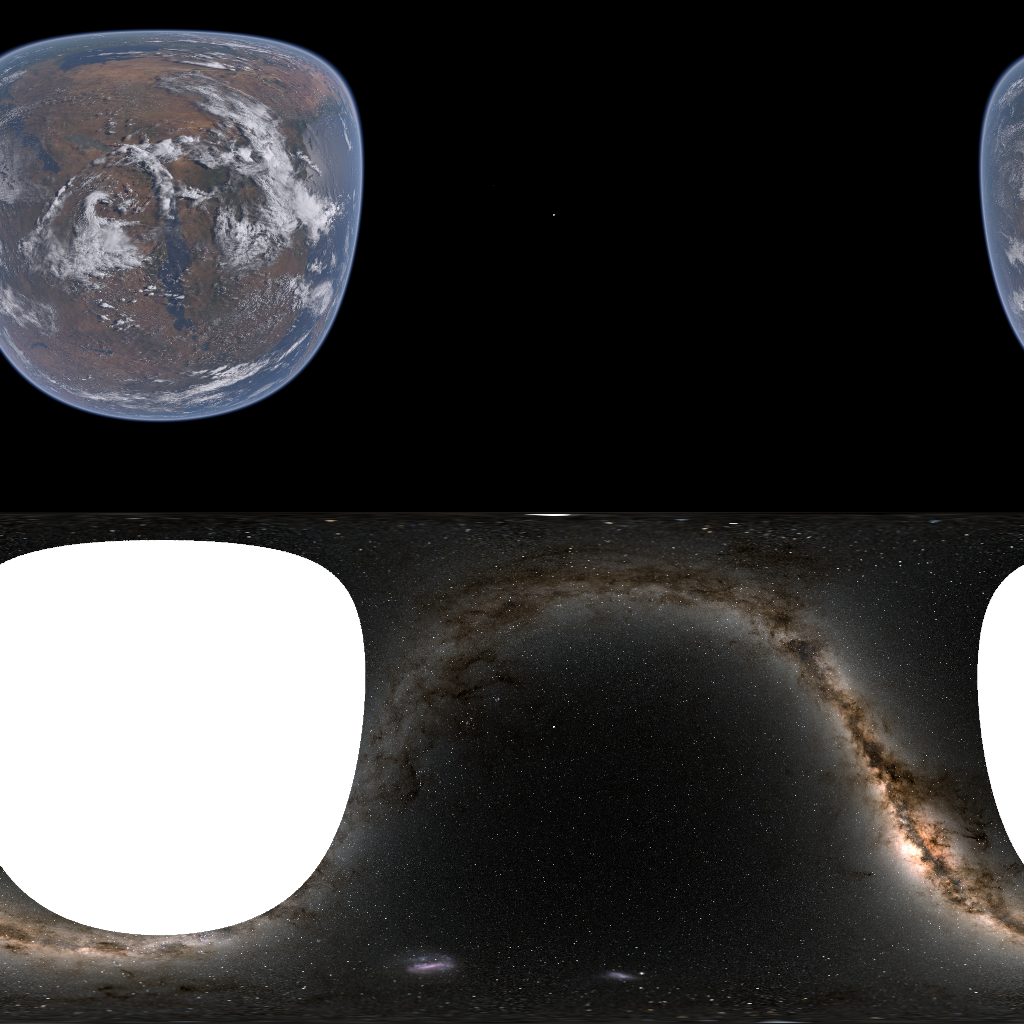}
    \caption{360-degree HDR environment maps}
    \label{fig:fig2}
\end{figure}

\subsection{Post-processing pipeline}
To further bridge the gap between synthetic and real-world imagery, the DLVS\textsuperscript3 framework incorporates a comprehensive post-processing pipeline developed using OpenCV. This pipeline applies a suite of visual effects and sensor artifacts to the rendered images, ensuring that the synthetic data more closely resembles the outputs of actual spaceborne cameras. The post-processing steps are designed to introduce controlled, physically plausible imperfections and characteristics commonly observed in real mission imagery.

The first steps involve converting the rendered images from a linear color space to the standard sRGB color space, mimicking compressed frames of camera sensors and display devices.

For applications that require monochromatic imagery, such as those using panchromatic sensors, the pipeline can simulate this effect by combining the color channels into a single grayscale image. This emulates the response of spaceborne cameras that operate without color filters.

To replicate the natural light falloff toward the edges (vignetting), the pipeline applies a radial gradient, resulting in darker corners and a brighter center.

The post-processing also models blooming and lens flares. Blooming occurs when bright light sources cause a soft halo or glow, while lens flares and streaks result from internal reflections and diffraction within the camera lens assembly. These effects are procedurally added to the images, introducing the kind of visual artifacts frequently seen in space imagery, especially when the Sun is present in the field of view.

Finally, the pipeline introduces camera noise, including both random (Gaussian) noise and photon shot noise (Poission), as well as occasional salt-and-pepper noise that simulates random pixel corruption.

By systematically applying these post-processing steps, the DLVS\textsuperscript3 dataset achieves a higher degree of visual realism and variability as seen in Figure \ref{fig:fig5}. This not only makes the data more suitable for training robust deep learning models, but also ensures that the models are better prepared to handle the diverse and imperfect conditions encountered in real-world space operations.

\section{The HST pose estimation dataset}
The HST pose estimation dataset\cite{dlvs3_hst_dataset_2025} is a demonstration cornerstone for advancing autonomous satellite navigation and was generated with DLVS\textsuperscript3.
The HST was chosen because it is one of the best documented and most visited satellite in the history of spaceflight, with thousands of photos available for visual inspection. However, it is important to note that very special photos were taken during the STS missions to capture as many details as possible: – the Space Shuttle, the HST, the position of the Earth, and the Sun are not random; during the documentation, astronauts tried to ensure homogeneous lighting conditions. The Sun rarely illuminates the HST directly; most of the time, it is covered by the Shuttle, whose large white surfaces reflect the Earth’s light at a wide angle.

The first part of the dataset contains 640.000 synthetic floating-point HDR multichannel images in OpenEXR format at a resolution of 1024×1024. The following data channels are included for each sample:

\begin{itemize}
\item Color image: 16-bit floating-point HDR (High Dynamic Range) image format, which allows each color channel to represent a much wider range of luminance values than standard 8-bit images. This increased precision is essential for accurately capturing the subtle variations in lighting and reflections present in space scenes, as well as for preserving highlight and shadow detail without clipping. This makes the data suitable for both advanced post-processing and high-fidelity deep learning applications.
\item Camera normal: This image encodes the surface normal vectors of the Hubble Space Telescope as seen from the camera’s perspective. Stored in a 16-bit floating-point RGB format, each pixel’s RGB values represent the X, Y, and Z components of the normal vector at that surface point in camera space (where the camera’s origin is at 0, 0, 0). This data is crucial for understanding the object’s orientation relative to the viewer.
\item Depth map: This grayscale image represents the distance of each visible point on the Hubble Space Telescope from the camera in meters. The data is stored in a 16-bit floating-point format, where each pixel’s intensity value corresponds to the depth at that pixel. This channel offers direct information about the spatial layout of the scene along the viewing direction.
\item Detailed mask: This grayscale image provides detailed semantic segmentation of the Hubble Space Telescope into its constituent parts. Stored as cryptomatte data embedded in EXR file assigns a unique 4-byte identifier to each pixel within a specific mask. The EXR header contains a manifest that links these identifiers to their corresponding mask names. In the case of the Hubble Space Telescope, this means 19 distinct and non-overlapping masks are defined, which collectively form the complete telescope.
\end{itemize}

\begin{figure}[H]
  \centering
  \includegraphics[width=0.9\columnwidth]{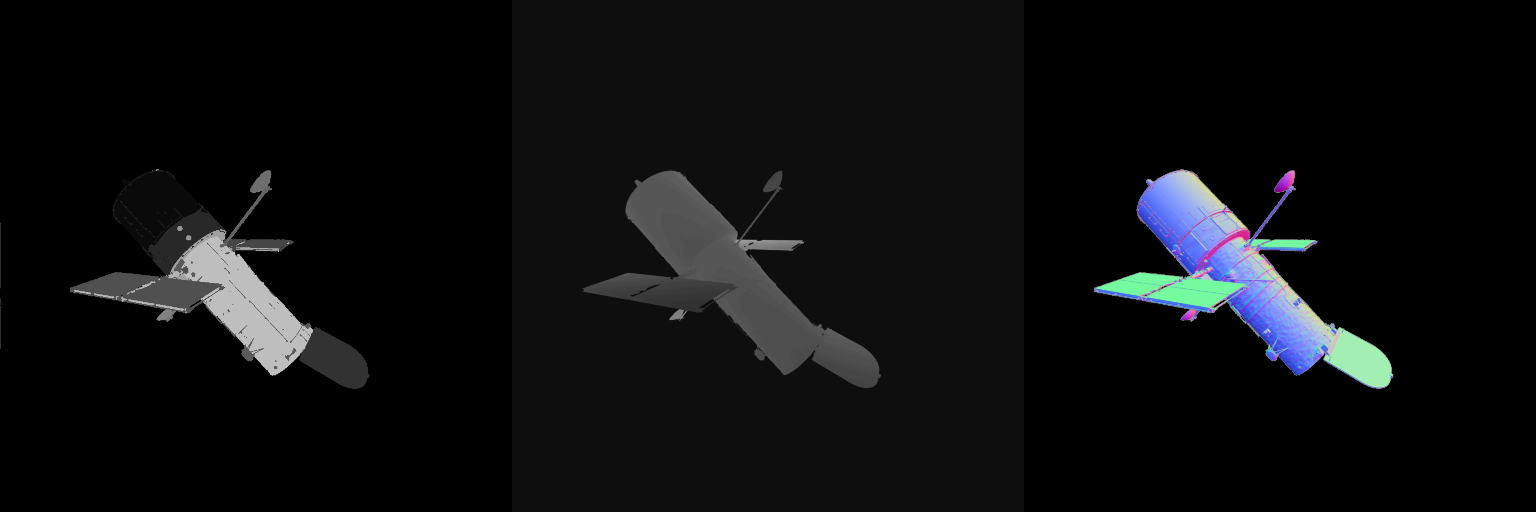}
  \caption{Different layers in the EXR image}
  \label{fig:fig3}
\end{figure}

\subsection{OpenEXR (.exr) format}
This is a versatile and powerful image format, particularly well-suited for storing high-dynamic-range (HDR) imagery and multiple image elements within a single file. A key advantage of the EXR format, as utilized by the DLVS\textsuperscript3 studio rendering pipeline, is its ability to embed various image buffers or “layers” within a single file. This means that for each rendered scene, the .exr file encapsulates more than just the final color image. It efficiently stores different image components.

Furthermore, the EXR format offers significant flexibility in terms of data representation for each of these components. Unlike conventional image formats with fixed channel counts and data types, EXR allows each embedded layer to have an arbitrary number of channels (e.g., single-channel grayscale for depth, three channels for RGB or normals) and to be stored with different numerical precisions.

\subsection{Keypoints}
The 3D model of the Hubble Space Telescope used in this dataset is annotated with a set of 37 distinct keypoints. These keypoints are strategically positioned on significant features and extremities of the model. While the current count is fixed at 37, it’s important to note that the number and placement of these keypoints can be adjusted in future iterations or for specific application requirements.

\begin{figure}[H]
  \centering
  \includegraphics[width=0.9\columnwidth]{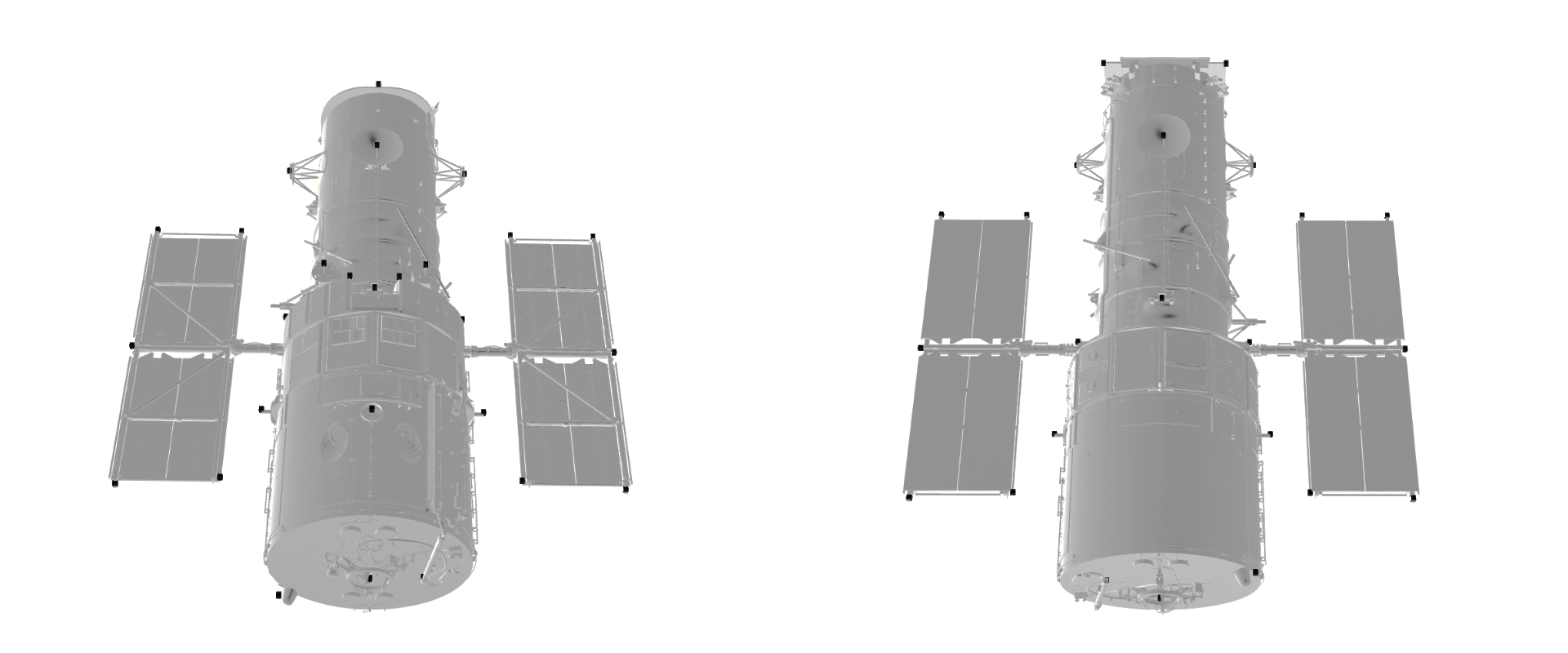}
  \caption{HST Keypoint positions}
  \label{fig:fig4}
\end{figure}

The primary purpose of these keypoints is to provide a sparse yet informative representation of the Hubble’s 3D pose and spatial extent. By tracking the 2D projections of these 3D keypoints in the rendered images, it becomes possible to estimate the telescope’s orientation and position relative to the camera. The distribution of these 37 keypoints is designed to capture the overall structure and silhouette of the Hubble Space Telescope effectively, ensuring that the entire model is well-defined and delimited within the 3D space and its 2D projections.

\subsection{Depth map}
The EXR contains the “depth” part, which stores the distance of each visible point on the Hubble Space Telescope from the camera in meters. In this dataset, the typical distance of the telescope from the camera ranges from 15 to 30 meters.

\subsection{Normal map}
DLVS\textsuperscript3 uses the standard normal map conventions:
Unit Normal vectors corresponding to the u,v texture coordinates are mapped onto normal maps. Only vectors pointing towards the viewer (z: 0 to -1 for left-handed orientation) are present, since the vectors on geometries pointing away from the viewer are never shown.

\subsection{Randomized materials} 
The material properties of the Hubble Space Telescope model were subtly altered in every image. This randomization of material characteristics further enhances the dataset’s variability, which is especially important because very little is known about the condition of materials used over a 25–30 year period.

Similarly, as the MLI wrinkles are unknown, a different state was procedurally rendered in each image. The reflectivity and roughness of the surfaces are also altered due to atomic oxygen and micrometeorite impacts.

The purpose of randomization is to systematically explore the event space around the unknown reality and thus provide a more robust solution. As a result, it is important to recognize that not every randomly selected image from the dataset is intended to be a perfect photorealistic match to actual space imagery. Instead, the emphasis is placed on generating a diverse range of plausible scenarios, so that models trained on this data can better adapt to the full spectrum of real-world variations that may be encountered. This is considered a working hypothesis, which is intended to be validated in future studies. Users of the dataset are also encouraged to contribute to this validation by sharing their findings and results regarding the effectiveness of domain randomization for generalization and robustness in their applications.

\subsection{Post-processing in the HST dataset}

\begin{figure}[H]
    \centering
    \includegraphics[width=0.9\columnwidth]{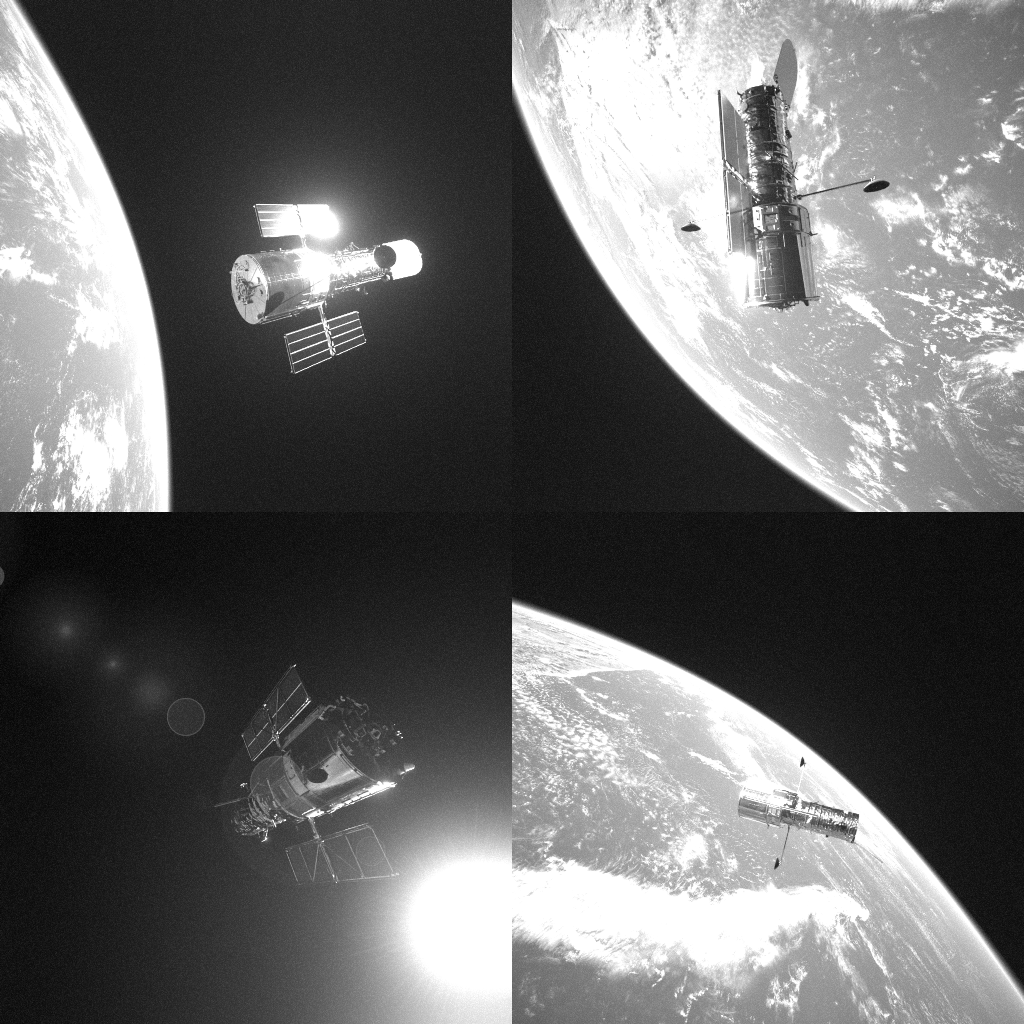}
    \caption{Post-processing on HST dataset images}
    \label{fig:fig5}
\end{figure}

As detailed in the general section on post-processing, the HST dataset includes a set of standard post-processing effects to enhance realism. Specifically, the following effects were applied: panchromatic filtering, vignetting, blooming, lens flaring, and Gaussian noise. The resulting images are provided in PNG format alongside the original EXR files. For users with specific knowledge of their target camera and sensor, it is advisable to develop a custom post-processing pipeline over the original EXR data tailored to specific requirements.

\subsection{Metadata}
The positions of the images in the database follow the true path of the Hubble Space Telescope starting at 06:46:00 UTC on 29/04/2025. A random time was selected from each 5-minute interval and images rendered at 100 perturbed chaser and target positions, where chaser is the active satellite or camera platform capturing images and performing pose estimation and the target is he observed satellite whose position and orientation are to be determined, in this case, the Hubble Space Telescope. Each 1000-image subset is composed of the daytime portion of one orbital period. The subsequent image sequence is initiated from the following dawn, resulting in 640 orbits being completed by the HST across the entire dataset.

Metadata information stored:
\begin{itemize}[itemsep=0pt, parsep=0pt]
\item FrameIndex (Int32)
\item ImagePath (String)
\item Epoch (Datetime)
\item SunPosition (Vector3)
\item EarthPosition (Vector3)
\item MoonPosition (Vector3)
\item ChaserPosition (Vector3)
\item TargetPosition (Vector3)
\item CameraTransform (Matrix4x4)
\item TargetTransform (Matrix4x4)
\item SunCenter (Vector2\footnote{Vector2 is in screen pixel coordinate system})
\item EarthCenter (Vector2)
\item MoonCenter (Vector2)
\item TargetCenter (Vector2)
\item TargetKeypoints (Array of 37x Vector2) 
\end{itemize}

\section{Conclusion and Future Work}
The DLVS\textsuperscript3 HST dataset represents a significant advancement in the field of synthetic data generation for satellite pose estimation. By leveraging advanced rendering technologies, physically accurate material modeling, and comprehensive domain randomization, the dataset provides a robust foundation for training and benchmarking deep learning models under a wide range of realistic and challenging visual conditions. 
\subsection{Initial release}
The first phase of the dataset consists of 640,000 high-fidelity images. To support broad adoption and experimentation, a demonstration subset (10,000 images + metadata, ~45 GB) is available via file sharing, while the full dataset (3.2 TB compressed) can be provided to interested parties over self-hosted SFTP. This initial release contains the rigid-body configuration of the HST, establishing a robust baseline for 6-DoF pose estimation tasks.

\subsection{Articulation: The next leap in complexity}
A major planned addition is the introduction of articulated parts, scheduled for release in 2025 Q3. This extension will contribute an additional 320,000 images in which key components of the HST - such as solar panels, antennas, and aperture door - are rendered in a variety of physically plausible configurations. The inclusion of articulated elements marks a fundamental shift: the HST is no longer treated as a simple rigid body, but as a complex system of interconnected, movable parts. This transition is essential, as real satellites possess joints and deployable mechanisms whose positions can vary independently, significantly complicating the pose estimation problem. With articulation, the dataset will move beyond the capabilities of standard 6DoF PnP solvers, requiring more advanced algorithms that can handle variable chains and part-level ambiguity.

\subsection{Scenario-based dataset}
The final planned extension is a 40,000-image scenario set, featuring 8-10 complete approach, flyby, and contact sequences. These curated trajectories are designed to support research into temporal and sequence-based pose estimation, as well as to provide benchmark scenarios for real mission planning and validation.

\end{multicols}

\begin{figure}[H]
    \centering
    \includegraphics[width=1.0\textwidth]{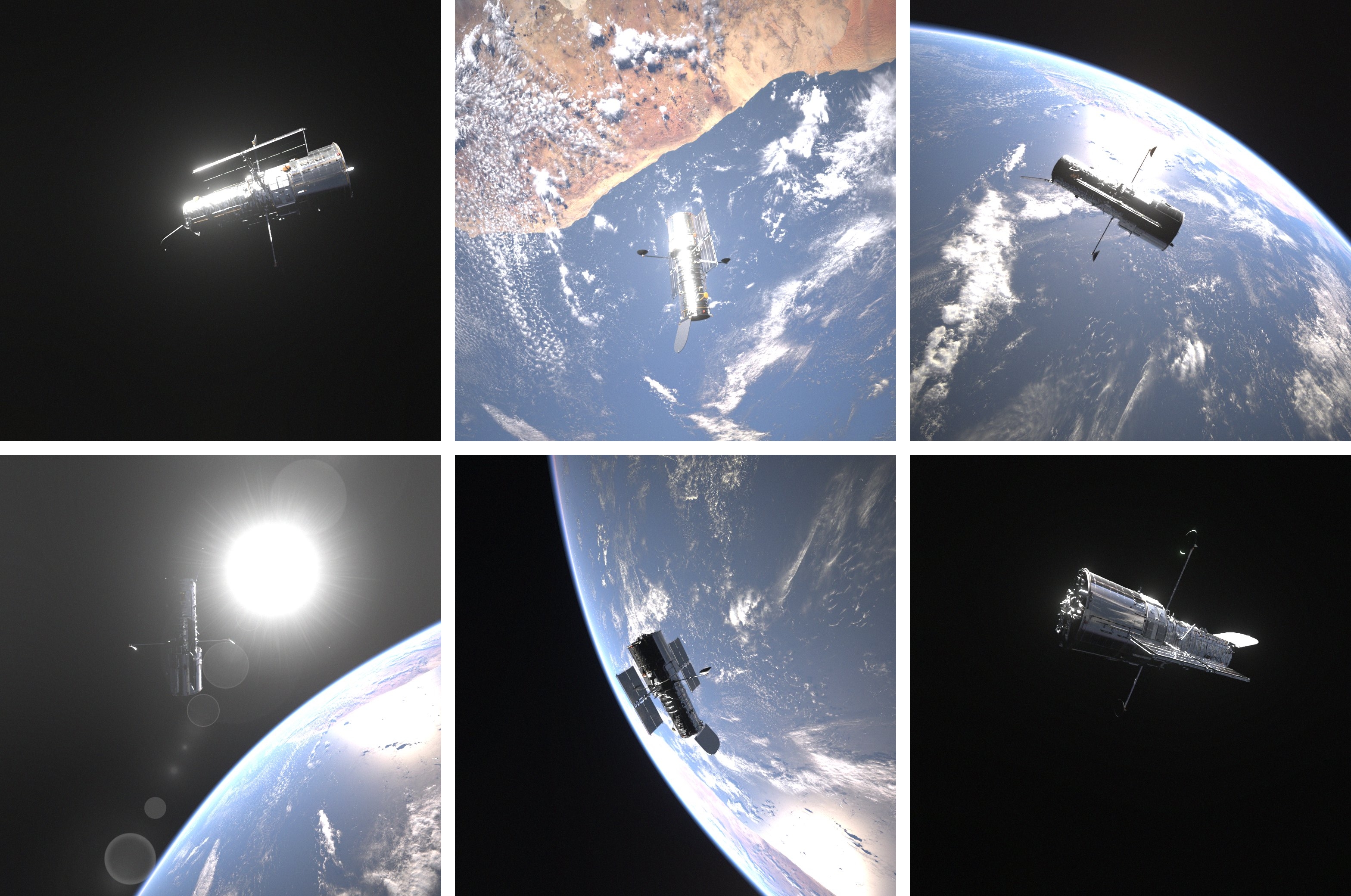}
    \caption{Full resolution examples from the dataset}
    \label{fig:fig6}
\end{figure}
\clearpage
\section{Quantitative dataset summary}
The following quantitative summary provides an overview of the main technical parameters and annotation types included in the DLVS\textsuperscript3-HST dataset. 
\begin{table*}[h]
\caption{Key quantitative and technical parameters of the DLVS\textsuperscript3 HST dataset.}
\centering
\begin{tabular}{ll}
\toprule
\textbf{Parameter} & \textbf{Value/Description} \\
\midrule
Total images (initial phase)          & 640,000 \\
Planned articulated images (Q3 2025)  & 320,000 \\
Scenario/trajectory images (final)    & 40,000 \\
Full planned dataset size             & 1,000,000 images \\
Image resolution                      & 1024 $\times$ 1024 pixels \\
Camera field of view (FOV)            & 90$^\circ$ (ideal pinhole camera, distortion-free) \\
Coordinate system                     & IAU\_Earth, shifted to target satellite center \\
File formats                          & EXR (raw, HDR, multi-layer), PNG (post-processed) \\
Annotation types                      & 2D keypoints (up to 37), segmentation, depth, normals, transformations \\
Satellite model detail                & $>$3.5 million polygons, cm-level accuracy \\
Rendering engines                     & Unreal Engine 5 (real-time), Houdini Karma (offline) \\
Environment map resolution            & Up to 4096 $\times$ 2048 px (HDRI) \\
Material library                      & Custom MaterialX, with aging and procedural variation \\
Articulated components                & Solar panels, antennas, doors, cables (bone animation, blend shapes) \\
Post-processing effects               & Panchromatic, vignetting, blooming, lens flare, Gaussian noise \\
Typical image size (PNG)              & $\sim$300--700 KB \\
Typical image size (EXR)              & $\sim$1--11 MB \\
Compressed full dataset size          & $\sim$3.2 TB \\
Download/demo subset                  & 10,000 images ($\sim$45 GB) \\
\bottomrule
\end{tabular}
\label{tab:hst_dataset_summary}
\end{table*}

DLVS\textsuperscript3 is distinguished by its scale, comprehensive annotation (including segmentation and depth), support for random and trajectory-based poses, variable lighting, and advanced simulation of lighting sources.

\begin{table*}[h]
\caption{Comparison of major public satellite pose estimation datasets}
\centering
\small
\begin{tabular}{lccccccccc}
\toprule
\textbf{Property} & \textbf{SPEED} & \textbf{SPEED+} & \textbf{SHIRT} & \textbf{SPADES} & \textbf{SEENIC} & \textbf{SPARK} & \textbf{URSO} & \textbf{DLVS3} \\
\midrule
Synthetic & \checkmark & \checkmark & \checkmark & \checkmark & \checkmark & \checkmark & \checkmark & \checkmark \\
Robotic Testbed & \checkmark & \checkmark & \checkmark & \checkmark & \checkmark & \checkmark &   &   \\
Event camera &   &   &   & \checkmark & \checkmark &   &   &   \\
RGB camera & \checkmark & \checkmark & \checkmark & \checkmark & \checkmark & \checkmark & \checkmark & \checkmark \\
Random poses & \checkmark & \checkmark &   & \checkmark &   & \checkmark & \checkmark & \checkmark \\
Trajectory &   &  & \checkmark &  & \checkmark &   &   & \checkmark \\
Relative pose & \checkmark & \checkmark & \checkmark & \checkmark & \checkmark &   & \checkmark & \checkmark \\
Keypoints & \checkmark & \checkmark & \checkmark &   &   &   &   & \checkmark \\
Celestial position &   &   &   &   &   &   &   & \checkmark \\
Depth map &   &   &   &   &   & \checkmark &   & \checkmark \\
Normal map &   &   &   &   &   &   &   & \checkmark \\
Segmentation &   &   &   &   &   & \checkmark &   & \checkmark \\
Nb. of images & ~15,000 & 70,000 & 4700 & 179,400 & 10,000 & 120,000 & 150,000 & \textbf{1,000,000} \\
Img. resolution & 1920x1200 & 1920x1200 & 1920x1200 & 1280x720 & 640x480 & 1440x1080 & 1024x1024 & 1024x1024 \\\bottomrule
\end{tabular}
\label{tab:dataset_comparison}
\end{table*}
\clearpage

\begin{multicols}{2}
\section{Dataset access}
The DLVS\textsuperscript3 HST dataset is available for academic and research purposes. For access to the full dataset (3.2 TB compressed), interested parties are invited to visit the official dataset portal\cite{dlvs3_hst_dataset_2025} or contact the authors.

\bibliographystyle{unsrt}  
\bibliography{references}  

\end{multicols}

\end{document}